# *Quality expectations of machine translation*

**Andy Way**

Adapt Centre, School of Computing, Dublin City University, Dublin, Ireland

**Abstract** Machine Translation (MT) is being deployed for a range of use-cases by millions of people on a daily basis. There should, therefore, be no doubt as to the utility of MT. However, not everyone is convinced that MT can be useful, especially as a productivity enhancer for human translators. In this chapter, I address this issue, describing how MT is currently deployed, how its output is evaluated and how this could be enhanced, especially as MT quality itself improves. Central to these issues is the acceptance that there is no longer a single 'gold standard' measure of quality, such that the situation in which MT is deployed needs to be borne in mind, especially with respect to the expected 'shelf-life' of the translation itself.

## 1 Machine Translation Today

Machine Translation (MT) is being deployed for a range of use-cases by millions of people on a daily basis. I will examine the reasons for this later in this chapter, but one inference is very clear: those people using MT in those use-cases *must* already be satisfied with the level of quality emanating from the MT systems they are deploying, otherwise they would stop using them.

That is not the same thing at all as saying that MT quality is perfect, far from it. The many companies and academic researchers who develop and deploy MT engines today continue to strive to improve the quality of the translations produced. This too is an implicit acceptance of the fact that the level of quality is sub-optimal – for some use-cases at least – and can be improved.

If MT system output is good enough for some areas of application, yet at the same time system developers are trying hard to improve the level of translations produced by their engines, then translation quality – whether produced by a machine or by a human – needs to be *measurable*.

Note that this applies also to translators who complain that MT quality is too poor to be used in their workflows; in order to decide that with some certainty – rather than rejecting MT out-of-hand merely as a knee-jerk reaction to the onset of this new technology – the impact of MT on translators' work needs to be measurable.

In Way (2013), I appealed to two concepts, which are revisited here, namely:

1. Fitness for purpose of translations,[1] and
2. Perishability of content.

In that work, I noted that:

> "the degree of human involvement required – or warranted – in a particular translation scenario will depend on the purpose, value and shelf-life of the content. More specifically, we assert that in all cases, the degree of post-editing or human input should be clearly correlated with the content lifespan."

In that paper, I also put forward the view that if there ever truly was a single notion of quality as regards translation – namely 'perfect' human translation – then this needs to be abandoned forthwith; the range of situations in which MT is being deployed nowadays includes many where there simply is no place for human intervention, either in terms of speed, or cost, or both.

In the remainder of this chapter, I will attempt to place MT in its proper place as we approach 2020. This involves examining the use-cases in which MT is deployed today, and how MT quality is measured. I will demonstrate how the construction of MT systems is changing, and describe ways in which MT evaluation needs to change. At all times, I will bear in mind the two constructs above as being of utmost importance when thinking about these issues: how will the translation be used, and for how long will we need to consult that translation?

## 2    Machine Translation Use Today

There are many estimates as to how much the translation industry is worth today, and how much it will expand over the coming years. For example, the size of the overall global language industry in 2015 was estimated at $38 billion, with estimates of up to $46 billion by 2016.[2] Thought leaders in this space have even begun to estimate the worth of the MT sector itself; in August 2014, TAUS stated that the MT industry was worth $250M.[3]

This was a significant announcement for a number of reasons. First and foremost, it recognised that MT was *already* being used successfully for a number of use-cases; secondly, it noted that while this estimate might be seen to be on the low side, for MT companies even a small slice of $250M was not to be sniffed at;[4] and thirdly, it pointed out that MT technology is a key enabler and a force multiplier for new

---

[1]   This concept is also applied to crowdsourced translation by Jimenéz-Crespo in this volume.
[2]   https://www.gala-global.org/industry/industry-facts-and-data
[3]   https://www.taus.net/think-tank/news/press-release/size-machine-translation-market-is-250-million-taus-publishes-new-market-report
[4]   Technavio estimate that the MT market will grow at a CAGR rate of 23.53% during 2015—19 (http://www.slideshare.net/technavio/global-machine-translation-market-20152019)

services, with innovative companies in IT and other sectors converging MT technology in new applications and products or using MT to enhance their existing products.

As we have pointed out before (Penkale and Way 2013; Way 2013), some translators like to pour scorn on the capability of MT, but by any measure, there is no real doubt that MT is being used at scale on a global basis every day. Back in 2012, Franz Och, who headed up the Google Translate team at that time, stated that:

> "Today we have more than 200 million monthly active users on translate.google.com. In a given day we translate roughly as much text as you'd find in 1 million books. To put it another way: what all the professional human translators in the world produce in a year, our system translates in roughly a single day."[5]

At a rough estimate, in 2012 Google was translating around 75 billion words per day. At the Google I/O event in May 2016,[6] Google stated that the average daily volume is about 143 billion words a day across 100 language combinations (see Figure 1), meaning that their translation volume has more or less doubled in just 4 years.

**Fig. 1** Daily Translation Usage in Google Translate (May 2016)

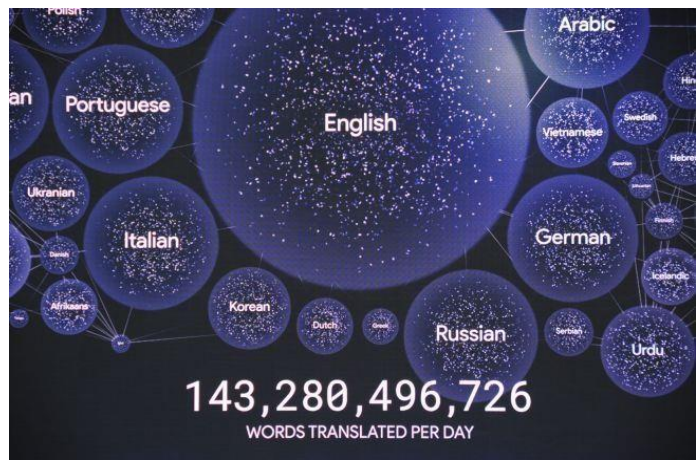

While Google translates by far the most words per day, other players also service huge amounts of translation requests. In March 2016, Joaquin Quiñonero Candela, Director of Engineering for Applied Machine Learning at Facebook, spoke about the amount of translation requests provided by his company today.[7] As shown in Figure 2, with 2 billion translations being provided on a daily basis, and almost 1 billion users seeing these translations each month, the numbers are truly staggering.

---

5  https://googleblog.blogspot.ie/2012/04/breaking-down-language-barriersix-years.html
6  https://events.google.com/io2016/

**Fig. 2** Translation Usage in Facebook (May 2016)

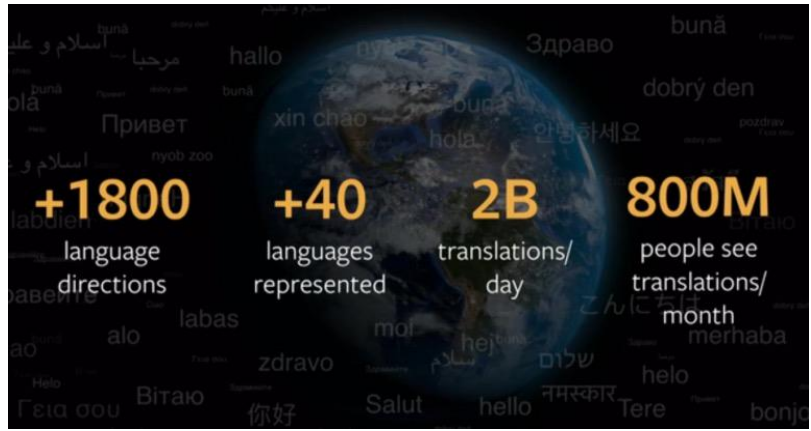

If all the translation requests that Bing Translator[8] and other online systems respond to on a daily basis are added in, this is a clear demonstration of the utility of online MT across a wide range of use-cases and language pairs to millions of distinct users.

Many other companies either produce generic MT toolkits available for purchase, or build customised engines that enable their clients to improve productivity, allow users to translate content previously not feasible due to time or cost constraints, and reduce time to market (see Way (2013) for a list of successful use-cases for MT).

Of course, there are other types of MT supplier too. One of these is KantanMT,[9] which like the Microsoft Translator Hub[10] allows users to upload their own translation assets and quickly build Statistical MT (SMT) systems with good translation performance in just a few hours. KantanMT have produced some impressive figures of their own in this space. Managing over 80 billion words of user-supplied MT engine training data, their platform currently performs 650 million translations each year.

Scale and robustness are one thing, but of course users care about quality too. Microsoft note on their webpage (see footnote 8) that "given the appropriate type and amount of training data it is not uncommon to expect gains between 5 and 10, even 15 in some instances, BLEU points on translation quality by using the Hub".[11]

---

[7] https://www.quora.com/Is-Facebooks-machine-translation-MT-based-on-principles-common-to-other-statistical-MT-systems-or-is-it-somehow-different
[8] https://www.bing.com/translator
[9] https://www.kantanmt.com/
[10] https://www.microsoft.com/en-us/translator/hub.aspx

This brings me to the main thrust of this chapter, namely how MT is evaluated, both in academic research labs and in industry, what is wrong with those methods, whether they are equally applicable to all MT system types, and how the myriad ways in which MT is being/will be deployed might affect the notion of utility in the future.

## 3   Machine Translation Evaluation Today

Of course, it is one thing building an MT system; it's another thing entirely knowing whether the quality produced is any good. In this section, I describe how MT quality has been measured over the years, focusing in particular on human and automatic evaluation, as well as task-based evaluation.

### 3.1  Human Evaluation

Human evaluation of MT quality goes back many years. There are different types of human evaluation of MT, including (Humphreys et al. 1991):[12]

- *Typological evaluation*, which addresses which translational phenomena can be handled by a particular MT system;
- *Declarative evaluation*, which addresses how an MT system performs relative to various dimensions of translation quality;
- *Operational evaluation*, which establishes how effective an MT system is likely to be (in terms of cost) as part of a given translation process.

---

[11]  In its original exposition in Papineni et al. (2002), the BLEU ("Bilingual Evaluation Understudy") score for a document was a figure between 0 and 1, the higher the better indicator of the quality of the MT system being evaluated. Here, and more commonly used nowadays in the field, this score is multiplied by 100 so that 'BLEU points' can be used to indicate progress compared to some benchmark.

[12]  We omit a lengthy discussion here on 'round trip' translation as an evaluation method (but cf. footnote 25), as it has been demonstrated by Somers (2005) to be an untrusted means of MT evaluation. In Way (2013), I note that in order to show that MT is error-prone, "sites like Translation Party (http://www.translationparty.com/) have been set up to demonstrate that continuous use of 'back translation' – that is, start with (say) an English sentence, translate it into (say) French, translate that output back into English, *ad nauseum* – ends up with a string that differs markedly from that which you started out with". I quickly show that such websites have the opposite effect, and observe that "It's easy to show MT to be useless; it's just as easy to show it to be useful, but some people don't want to".

Typological evaluations were conducted by MT system developers to ensure that their engines were continuing to improve over time. These were typically carried out with reference to a test suite of examples (King and Falkedal 1990; Arnold et al. 1993; Balkan et al. 1994). In particular, test suites need to be designed to ensure wide coverage of the source language as well as test certain key translational phenomena. King and Falkedal (1990) note that if this is to be done properly, "this assumes the availability of someone with at least a knowledge of the languages concerned, of linguistics and preferable some experience of machine translation". They quickly lament that this is rarely the case in practice.

One of the earliest, most well-known instances of a declarative (or "static") evaluation was the ALPAC report (Pierce et al. 1966). Translation quality was measured along two dimensions: (i) *fidelity* (or "informativeness"), the extent to which a translated text contains the same information as the source text; and (ii) *intelligibility*, the extent to which the output sentence is a well-formed example of the target language. In their study, a group of 18 English monolinguals and 18 English native speakers with a "high degree of competence in the comprehension of scientific Russian" were asked to evaluate 3 human and 3 machine translations on 9- and 10-point scales for the dimensions of intelligibility and informativeness, respectively.

In subsequent similar evaluations, more coarse-grained scales (with 1-4 or 1-5 rather than a 10-point scale) are typically used, as it is difficult for humans to discriminate with confidence using such a fine-grained scale. Note too that using fewer decision points tends to ensure greater inter-rater scoring consistency. In addition, terms such as "accuracy" (or "adequacy") and "fluency" are more likely to be seen as replacements for the terminology introduced by Pierce et al. (1966).

Arnold et al. (1994) observe that "accuracy testing follows intelligibility rating", but clearly the two evaluations are related; if a particular output string is deemed to be unintelligible (or disfluent), it is arguable whether there is any point in performing an accuracy test. When comparing the two tasks, Arnold et al. (*ibid*.) note that "accuracy scores are much less interesting than intelligibility scores … because accuracy scores are often closely related to the intelligibility scores; high intelligibility normally means high accuracy".[13] Until quite recently, it would have been rare indeed for an MT system to output a well-formed target translation that bore no resemblance to the input string, but as I note in Way (2018):

> "[Neural] MT output can be deceptively fluent; sometimes perfect target-language sentences are output, and less thorough translators and proofreaders may be seduced into accepting such translations, despite the fact that such translations may not be an actual translation of the source sentence at hand at all!"

Whatever terms are used, outputting numbers on scales as an assessment of quality gives the user little idea about the actual amount of effort it would take for the MT output to be post-edited into the final translation. Accordingly, operational evaluations were designed to take other factors into account, not just the quality of the MT output itself.

---

[13]     Indeed, the results from the ALPAC evaluation demonstrated there to be considerable correlation between intelligibility and fidelity.

Vasconcellos (1989) details one of the first studies comparing post-edited MT against human translation *per se*. While the findings were positive as far as MT usage at the Pan-American Health Organisation were concerned in terms of quality and speed, it did not attempt to measure the overall cost.

Humphreys et al. (1991) were one of the first to detail the contents of such an operational evaluation, addressing questions such as translator selection, quality required, how quality is to be assessed (e.g. scores according to scales, or ranking), speed improvements over time, type of data to be used, dictionary usage etc. In closing, they note that "an adequate model of operational evaluation requires very substantial input (in terms of subject numbers and subject and experimenter time)".

In concluding this section, it is worth pointing out that this remains a problem for the industry as a whole, with no clearly defined process as to whether MT should or should not be introduced into a company's translation workflow. Companies often overlook how disruptive a technology MT actually is: it impacts not just technically trained staff, but also project managers, sales and marketing, the training team, finance employees, and of course post-editors and quality reviewers. All of this should be taken on board beforehand if the correct decision is to be taken with full knowledge of the expected return on investment, but in practice it rarely is.

### 3.2 Automatic Evaluation

Despite its obvious benefits, human evaluation is slow, expensive and inconsistent. Metrics such as Word Error Rate (WER: Levenshtein 1966), and Position-Independent Word Error Rate (PER: Tillmann et al. 1997) had been used for some time to measure the effectiveness of automatic speech recognition (ASR). These methods simply examine how many of the target words are correct (PER) and in the right position (WER) compared to some human reference. This is fine for ASR, which is a monolingual task. These metrics have been used for MT evaluation, but of course, the effectiveness of the translation needs to be measured not just as an example target string ("fluency", in human terms), but also with respect to its being an accurate translation of the source ("adequacy"). For an overview of automatic MT evaluation metrics, see Castilho et al. in this volume.

#### 3.2.1 Inherent Problems with Automatic Evaluation Metrics

Nonetheless, when new metrics came in at the start of the century designed specifically for MT evaluation, they ignored the source sentence altogether. BLEU and NIST (Doddington 2002) both came on the scene at around the same time, and used different (but related) ways to compute the similarity between a human supplied 'gold standard' reference and the MT output string based (largely) on *n*-gram co-occurrence, i.e. how often words and phrases (up to length 4) occur both in the human reference and the MT output string.

As well as ignoring the source sentence altogether, a further problem for these evaluation metrics is that while they do function (better) with multiple human-supplied references (e.g. NIST MT-06 English-to-Chinese and English-to-Arabic tasks, which "had four independently generated high quality translations that were produced by professional translation companies"),[14] most of the time only a single reference is supplied; traditionally MT system developers hold out sections of the parallel training data – a (large) collection of source sentences and their translations, such as appears in a Translation Memory (TM: Heyn 1998) – to act as development (to optimally tune the system parameters)[15] and test sets.

There are a number of problems with this. One is that, as He and Way (2009a) demonstrated, Minimum Error Rate Training (Och 2003) – the tool most used for parameter tuning in SMT – is sub-optimal despite being tuned on translation-quality measures such as (document-level) BLEU.[16] More specifically, we showed that tuning on a particular objective measure (on the development set) cannot be guaranteed to deliver the optimal score on the test set, i.e. in order to deliver (say) the best BLEU score in testing, one might be better off tuning on (say) METEOR (Banerjee and Lavie 2005) rather than BLEU itself, as might be expected.

More importantly, as any pair of translators will tell you, there is no such thing as *the* correct translation. In a discussion regarding translator resistance towards MT in Way (2012), I summoned Bellos' (2011) observation that translators are often less than complimentary regarding each other's translations; specifically, he states that "translation commentators lead the field in throwing most of its work in the direction of the garbage dump", and soon thereafter that "it seems implausible that anyone would ever make such a statement about any other human skill or trade". Note that this can be observed in practice, too; when translators are asked to post-edit MT, they often make unnecessary changes, as while the MT output might have been acceptable to some, 'it's not quite how they would have said it themselves'. Note too the place of proofreaders in the human translation cycle, who alongside fixing errors made by the original translator, may be incentivised to make unnecessary changes to continue to justify their own positions in the workflow. In this regard, de Almeida (2013) observes that for both English-to-French and English-to-Spanish, 'essential' changes (language errors and mistranslations) comprise only about half of the overall edits made to documents, with the others concerning lexical choice, adding extra words, reordering, and even changing punctuation.

---

[14]   http://www.itl.nist.gov/iad/mig/tests/mt/2006/doc/mt06eval_official_results.html

[15]   Minimally, in an SMT system these would be the "translation model" inferred from the parallel data, which essentially suggests which target-language words and phrases might best be used to try to create a translation of the source string; and the "language model" inferred from large collections of monolingual data, and used to try to create the most likely target-language ordering of those suggested target words and phrases.

[16]   See Section 3.2.2 for discussion of document-level versus sentence-level MT evaluation.

Finally, here, as I noted in Way (2013), "MT developers are forced to (wrongly) assume human translations to be perfect when conducting automatic MT evaluation". As Penkale and Way (2013) observe, some translators still argue that there is only one level of quality – 'perfect' human translation – despite the myriad of use-cases available today, many of which omit a human in the loop entirely. Even with proofreaders in the fully managed translation, editing, and proofreading cycle, mistakes do occur, and sometimes these wrong human translations are precisely those against which the output from MT systems is compared against.

### 3.2.2 Problems with Automatic Evaluation Use

A further problem with the use of such metrics in practice is that BLEU is often used at the sentence level, either by system developers against a common test set to track system improvement over time, or in ranking tasks with human evaluators to try to seek insight into which translation produced by different systems might be 'better'. Of course, this is not the fault of the designers of the automatic evaluation metrics, which were designed to work at the document level. As a consequence, several variants of BLEU have been designed and can be used for that purpose (e.g. Lin and Och 2004; Liang et al. 2006).

Note too that He and Way (2009a, 2009b) demonstrate that certain automatic metrics prefer shorter/longer translations. METEOR prefers longer outputs than the reference translations owing to the different chunk penalties assigned to different languages.[17] He and Way demonstrate clearly that by imposing a static chunk penalty when tuning with METEOR gives better translation results when measured by BLEU, Translation Edit Rate (TER: Snover et al. 2006) and METEOR.

TER, in contrast, prefers shorter sentences to be generated by MT. As it is an error metric, the more words generated by an MT system, the larger the number of insertions, deletions and substitutions which will typically be required to transform the MT hypothesis into the reference sentence. He and Way (2009a) note that this is "less likely to harm in MT evaluation, unless a system is developed specifically to game the metric. However, if such knowledge is made use of in tuning, the system will be *tuned* to take advantage of this preference, and will tend to output overly succinct sentences" (original emphasis).

---

[17] METEOR rewards MT output composed of fewer chunks. Output containing bigram (or longer) matches compared to the reference translation are penalised less than those comprising unigram matches only.

### *3.2.3 Does Automatic Evaluation Corroborate Human Evaluation?*

If automatic evaluation metrics are to be of any use as arbiters of translation quality, then the predictions they make should correlate with human judgement, assuming humans can be trusted to evaluate MT output in a consistent fashion.

Since its introduction over 15 years ago, BLEU has been by some distance the most reported metric in papers involving MT experiments. While early studies (Doddington 2002; Coughlin 2003) demonstrated its correlation with human judgements of translation under certain circumstances, it has been widely accepted for some time now that BLEU has many limitations.[18]

Callison-Burch et al. (2006) explain that BLEU places no explicit constraints on the order in which matching *n*-grams occur, in order to permit variation in word choice in MT outputs. In the multiple reference scenario, matches can be extracted in a huge number of different ways, so that millions of variants all receive the same BLEU score for any particular translation hypothesis. They argue that "because the number of translations that score the same is so large, it is unlikely that all of them will be judged to be identical in quality by human annotators". By extension, they note that translations with higher BLEU scores might be deemed worse by human judges.

Hovy and Ravichandran (2003) give a nice example to illustrate this point. Let us assume the reference translation to be (1):

(1) The President frequently makes his vacation in Crawford Texas .
(2) George Bush often takes a holiday in Crawford Texas .
(3) holiday often Bush a takes George in Crawford Texas .

If the two MT hypotheses in (2) and (3) were produced, then they would receive exactly the same BLEU score! Why? Firstly, both (2) and (3) are the same length, so the brevity penalty[19] plays no role here. They both share the 4-gram "in Crawford Texas .", the trigrams "in Crawford Texas" and "Crawford Texas .", the bigrams "in Crawford", "Crawford Texas", and "Texas .", and the unigrams "in", "Crawford", "Texas", and ".". All the other words in (2) and (3) are treated by BLEU as non-matches; there is no benefit gained by the fact that we know – at least in 2003! – the phrase "George Bush often takes a holiday" is synonymous with "The President frequently makes his vacation" in (1). As Babych and Hartley (2004) pointed out, BLEU weights all items in the reference sentence equally, so the fact that for the most part (3) is word salad makes no difference to its overall BLEU score.

---

[18] Nonetheless, more recent papers (Agarwal and Lavie 2008; Farrús et al, 2012) have also demonstrated that BLEU correlates extremely well with human judgement of translation quality.

[19] This was introduced to prevent systems from outputting very short target-language strings (such as "the") but nonetheless obtaining a high score. Accordingly, the shorter the translation compared to the reference translation, the more punitive the brevity penalty.

Note too that BLEU pays no attention to semantic errors either. This is easily demonstrated with the (somewhat artificial) example in (4):

(4) George rhododendron often takes a holiday in Crawford Texas .

That is, assuming (again) the reference translation to be (1), the BLEU score for (2) and (4) would again be identical; despite the fact that rhododendron is a type of bush, it is clearly an inferior translation compared to (2). To BLEU, both "Bush" (in (2)) and "rhododendron" (in (4)) are simply words that do not occur in the reference (1), so are treated exactly the same.

As a result of these sorts of problems, diagnostic MT evaluation emerged as a sub-field in its own right, as a means of capturing more precisely the types of errors made by various systems.[20] Vilar et al. (2006) observe that "a relationship between [automatic] error measures and the actual errors found in the translations is … not easy to find". Accordingly, they produced a human error analysis and error-classification scheme which extends the error typology presented in Llitjós et al. (2005), as a means of focusing research effort. Popović and Ney (2011) present the first steps towards a framework for automatic analysis and classification of errors, while Naskar et al. (2011) produced the DELiC4MT system, which identified user-specified fine-grained classes of translation errors based on the linguistic shortcomings of the particular MT system. This is of interest to users to gain insights into the linguistic strengths and weaknesses of the MT system, but also allows the MT system developers to try to correct these errors and improve translation performance.

Returning to the merits of the automatic MT metrics *per se*, Callison-Burch et al. (2006) go on to explain why human rankings of translation systems do not tally with automatic rankings computed via the BLEU score. Firstly, as I pointed out at the beginning of Section 3.2.1, humans are asked to evaluate fluency without recourse to the source sentence; while this can of course be done as a monolingual task, in reality a human would surely calculate fluency and adequacy at the same time. Secondly, the authors demonstrate an inherent bias in BLEU (and similar metrics) against systems (such as rule-based systems) which do not have at their core a component founded on $n$-gram statistics.[21]

Callison-Burch et al. (2008) show that that BLEU has a lower correlation with human judgement than metrics which take into account linguistic resources (such as part-of-speech tags) and better matching strategies, e.g. METEOR computes word matching based on stemming and WordNet synonymy (Miller et al. 1990).

A recent paper by Smith et al. (2016) serves to remind us of the fallibility of BLEU. They note that it "can be 'cheated': very bad translations can get high BLEU scores", although they do accept that their experiments "used BLEU in a very different fashion from that for which it was designed" (*op cit*.).

---

[20] See Popović (this volume) for a discussion of the evolution of diagnostic MT error typologies.
[21] 'Phrases' in phrase-based SMT refer only to $n$-gram sequences, i.e. contiguous sequences of surface words, not to the linguistic "constituent" sense of the word.

In the interim, partly because the community knows that better evaluation measures are needed, many other types of MT metrics have been developed which exploit deeper features such as paraphrases (Zhou et al. 2006), or syntax (Liu and Gildea 2005; Owczarzak et al. 2007), as well as metrics that try to exploit machine-learning techniques (Albrecht and Hwa 2007; Ye et al. 2007; He and Way 2009c). The Rank-based Intuitive Bilingual Evaluation Score (RIBES: Isozaki et al. 2010) was developed especially to take reordering into account between languages with very different word orders, and has been demonstrated to have high correlation with human evaluations of MT systems.

Nonetheless, none of these metrics are widely used. Despite the well-known problems with BLEU, and the availability of many other – arguably better – metrics, MT system developers have continued to use it in the intervening 10 years as the primary measure of translation quality, in academic circles especially.

### *3.3 Task-Based Evaluation*

When evaluation of MT systems started to be taken seriously in the 1990s, some of the early papers on the topic (e.g. Doyon et al. 1999) noted that the aims which the technology was expected to be used for had to be known in advance. Of course, some objectives could be more tolerant of MT errors than others.

Taking this on board, more and more evaluations have taken place in the interim with the specific task in mind (e.g. Thomas (1999) for spoken-language MT; or Voss and Tate (2006) for information extraction). Indeed, WMT evaluations[22] regularly include specific tasks nowadays, including medical translation (e.g. Zhang et al. 2014), automatic post-editing (e.g. Chatterjee et al. 2015) and MT for the IT domain (e.g. Cuong et al. 2016). I take this as evidence that the community as a whole is well aware of the fact that when evaluating MT quality, the actual use-case and utility of the translations therein need to be borne in mind.

## 4    The Changing Nature of MT System Design

When MT evaluation began in earnest, most systems were rule-based (RBMT). Since then, of course, we have seen the advent of SMT, and the rise of automatic MT evaluation metrics, as described in the previous section. When these came in, many people noted that as they were largely *n*-gram-based, there was an implicit bias against RBMT, the output of which was demonstrated to be considerably better

---

[22]    The Workshop (now Conference) on Machine Translation runs annual competitive MT system evaluations for a range of tasks. See http://www.statmt.org/wmt17/ for the latest in the series.

in human evaluations (e.g. Riezler and Maxwell 2005; Farrús et al. 2012; Lewis and Quirk 2013).[23]

More recently, Neural MT (NMT, e.g. Sennrich et al. 2016a) has been demonstrated to be very competitive compared to state-of-the-art SMT models, albeit to date for a limited number of language pairs and document types. On average, improvements over Phrase-Based SMT (PB-SMT, cf. Koehn et al. 2003) were of the order of two BLEU points. Indeed, most evaluation of NMT was conducted in terms of automatic MT metrics until Jean et al. (2015) set a manual ranking evaluation with non-professional annotators.

As a result of this good performance, Bentivogli et al. (2016) undertook an indepth (diagnostic) human evaluation to try to understand exactly where these improvements in terms of automatic metrics actually came from. In particular, they leveraged high-quality post-edits performed by professional translators on 600 output sentences from both PB-SMT and NMT systems for English-to-German translation of TED talks. The NMT system of Luong and Manning (2015) was compared against a standard PB-SMT system (Ha et al. 2015), a hierarchical SMT system (Jehl et al. 2015) and a system combining PB-SMT and syntax-based SMT (Huck and Birch 2015). HTER – essentially, TER with a human-in-the-loop (Snover et al. 2006) – and multi-reference TER was used, and results showed that NMT outperformed the other approaches in all metrics at a statistical significance level of $p=0.01$.

From a linguistic point of view, NMT was seen to produce significantly fewer morphological errors (-19%), lexical errors (-17%), and substantially fewer word order errors (-50%) than its closest statistical competitor. With respect to word order, NMT demonstrated a 70% reduction in the incorrect placement of verbs, and an almost 50% reduction for erroneous noun placement. With respect to overall post-editing effort, NMT generated outputs that required about a quarter fewer edits compared to the best PB-SMT system.

However, while the body of evidence in favour of NMT continues to grow, it is unclear that NMT is in a position to replace SMT entirely just yet. For example, Bentivogli et al. (2016) found that NMT degrades with sentence length for transcribed speeches. While NMT outperformed SMT for subsets of all lengths in their dataset, the gap became smaller as sentence-length increased. Our own in-house tests have shown that for small amounts of good quality training data, NMT cannot outperform PB-SMT systems. Note too that NMT currently takes much more time to train, and translation speed is slower compared to PB-SMT.

Nonetheless, the translational improvements discovered by Bentivogli et al. (2016) lead me to think that *n*-gram-based metrics such as BLEU are insufficient to truly demonstrate the benefits of NMT over PB-SMT. This is especially the case as character-based models (Chung et al. 2016) – or combinations of word- and character-based models (e.g. Luong and Manning 2016) – become more prevalent, in which case evaluation metrics such as ChrF (Popović 2015) which operate at the character level become more appropriate.

---

[23] Over the past ten years or so, SMT system developers have been incorporating more and more linguistic features. It is interesting to ponder whether BLEU (and similar metrics) disadvantages such linguistically enhanced systems compared to 'pure' SMT engines, in much the same way as RBMT output was penalised compared to pure *n*-gram-based systems.

In practice, a 2-point improvement in BLEU score which was typically seen in WMT-2016 would be far too small to be noticed in a real industrial relevant translation task. If word order is drastically improved, and fewer morphological and lexical errors are being made in NMT, one would expect to see a *huge* improvement in terms of automatic evaluation metrics, rather than a relatively modest – albeit statistically significant – one.[24] If NMT does become the new state-of-the-art as the field expects, one can anticipate that further new evaluation metrics tuned more precisely to this paradigm will appear sooner rather than later.[25]

## 5     A View on Future Deployment of MT, and its Impact on Translation Evaluation

I began this chapter by emphasising that MT needs to be evaluated in the context of the use-case for which it is intended. Despite the ever-increasing range of use-cases that are springing up, one can predict with confidence that for many use-cases, MT output will continue to require post-editing by expert human translators.

Clearly, human translators have for some time now been using Translation Memory (TM) systems to good effect. TM systems work as follows (Moorkens and Way 2016):

> "TM systems search the source side of a set of translation pairs for the closest-matching instances above some pre-determined threshold imposed by the translator (so-called 'fuzzy matches'; Sikes (2007)). A ranked list of the said translation pairs is then presented to the translator with user-friendly colour-coding to help the user decide which parts are useful in the composition of the target translation, and which should be ignored and discarded."

Accordingly, translators have been accustomed to using fuzzy match score as a predictor of translation quality. What is more, translators can configure fuzzy match

---

[24]   Note, however, that the NMT system of Luong and Manning (2015) was more than 5 BLEU points better than a range of SMT systems for English to German. This sort of difference in BLEU score is more like what we might expect given the huge improvements in quality noted by Bentivogli et al (2016) in their study. In this regard, both Shterionov et al (2018) and Way (2018) note that BLEU may be under-reporting the difference in quality seen when using NMT systems, with the former attempting to measure the level of under-reporting using a set of novel metrics.

[25]   Without further comment, we merely note here that the 'round trip' (or 'back') translation discredited by Somers (2005) – cf. footnote 12 – has been demonstrated to be very useful in NMT as a means of generating additional 'synthetic' parallel training material (e.g. Sennrich et al. 2016b).

thresholds – a balance between precision and recall – themselves, so they remain in control of the translation process. In real translation pipelines, MT usually kicks in for matches below the fuzzy match threshold set by the user, despite the fact that many researchers (e.g. Simard and Isabelle 2009; Moorkens and Way 2016) have demonstrated that translator performance can be harmed by the imposition of such arbitrary cut-offs; MT can be better than TM above the fuzzy match threshold, while TM may have more utility than MT below it.

Nonetheless, recognising that translators will remain a large cohort of MT users compels MT developers to begin to output translations from their MT systems with an accompanying estimation of quality that makes sense to translators;[26] while BLEU score is undoubtedly of use to MT developers, outputting a target sentence with a BLEU score of (say) 0.435 is pretty meaningless to a translator. The automatic MT metric that is most appealing in this regard is TER, as it is indicative of the amount of post-editing (in terms of substitutions, insertions and deletions) required to produce a good quality target-language sentence from the MT output. In a few cases, where MT is integrated with TM, MT matches are output as 'just another match', with MT matches used to reinforce TM fuzzy matching (Hofmann 2015).

Translators are used to being paid different rates depending on the level of fuzzy match suggested by the TM system for each input string. With that in mind, another area where MT can work to the benefit of the translation community is in promoting fuzzy matches to the next highest level (Biçici and Dymetman 2008). Moorkens and Way (2016) also observe that translators are used to not receiving help from TM for all input sentences, so that "MT developers have allowed the soft underbellies of their engines to be exposed 'warts and all' to translators, as MT outputs are typically provided for every source segment". In order to prevent this, they suggest that better system-internal confidence measures are needed so that translators can learn to trust the MT output they are confronted with.

Finally, despite the fact that more and more use-cases are emerging where MT can be useful, and regardless of whether PB-SMT or NMT systems prevail, we can expect TM technology to remain as an essential tool in the translator's armoury. Many researchers have demonstrated how the two technologies can co-exist to good effect, either via system recommendation (He et al. 2010a,b) or by using fragments from TM in SMT (e.g. Koehn and Senellart 2010; Ma et al. 2011; Wang et al. 2013; Li et al. 2014). Whatever the individual set-up preferred by translators, candidate translation outputs have to come with a readily intelligible indicator of quality, lest their decision-making process be cognitively overloaded such that rather than being a translation aid, such tools turn out to actually be an impediment to improved translation throughput.

---

[26] The subfield of quality estimation (see Specia and Shah in this volume) attempts to predict whether a new source string will result in a good or bad translation. This is different from MT evaluation, where we have a reference translation to compare the MT hypothesis against *post hoc*.

## 6     Final Remarks

This chapter has addressed the notion of what level of quality can be expected from MT. MT is not going away; year on year, its usage is increasing exponentially, which is a clear indication that MT quality is continually improving. Accordingly, those translators who remain opposed to the improvements that can be brought about by MT are only hurting themselves.

At the same time, although MT quality is getting better all the time, for this to be truly impactful in industry, we will need to see a quantum leap in terms of improved output as measured by traditional automatic evaluation metrics, or where quality of newer systems is better reflected by more suitable novel metrics. While incremental improvements are to be welcomed, there are many more pressing concerns for industry, including better terminology integration, improvements in post-editing environments, and indeed novel pricing models.

Furthermore, there are use-cases emerging where there is no role for the human translator/post-editor, and where translation quality can only be interpreted in terms of fitness for purpose of the translation outputs. Nonetheless, many use-cases will continue to require humans to post-edit the translations output by MT systems. Accordingly, it behoves the entire MT developer community to deliver MT output with a score that is meaningful to human post-editors, so that they can immediately decide whether it is either quicker to post-edit the MT suggestion, or to translate the source string from scratch by hand.

The MT developer community continues to use automatic metrics most suited to evaluating MT output emanating from word- and phrase-based systems. In much the same way as metrics like BLEU were not suited to output coming from grammar-based systems, in this chapter, I hypothesise that they are not discriminative enough to accurately reflect the translation quality of (largely) character-based NMT systems. Accordingly, despite the fact that they are expensive to set up and slow to analyse the results, human evaluation of MT output remains crucial if system developers are to improve their systems still further.

**Acknowledgments** This work has been supported by the ADAPT Centre for Digital Content Technology which is funded under the SFI Research Centres Programme (Grant 13/RC/2106) and is co-funded under the European Regional Development Fund.